\documentclass{article}
\usepackage{amssymb}
\usepackage{graphicx}
\usepackage{amsmath}
\usepackage{booktabs}
\usepackage{tabularx}
\usepackage{float}
\usepackage{booktabs}
\usepackage{listings}
\usepackage{xcolor}
\usepackage{array}
\usepackage{caption}
\definecolor{codegray}{rgb}{0.5,0.5,0.5}
\definecolor{codepurple}{rgb}{0.58,0,0.82}
\definecolor{backcolor}{rgb}{0.97,0.97,0.97}

\lstdefinestyle{pythonstyle}{
    backgroundcolor=\color{backcolor},
    commentstyle=\color{gray}\ttfamily,
    keywordstyle=\color{blue},
    numberstyle=\tiny\color{codegray},
    stringstyle=\color{codepurple},
    basicstyle=\ttfamily\footnotesize,
    breaklines=true,
    frame=single,
    captionpos=b,
    keepspaces=true,
    numbers=left,
    numbersep=5pt,
    showspaces=false,
    showstringspaces=false,
    showtabs=false,
    language=Python
}

\usepackage[preprint]{corl_2025} % Uncomment for the camera-ready ``final'' version.

\title{CaseEdit: Enhancing Localized Commonsense Reasoning via Null-Space Constrained Knowledge Editing in Small Parameter Language Models.}

\author{
  Varun Reddy\textsuperscript{1}, \quad Yen-Ling Kuo\textsuperscript{1} \\
  \textsuperscript{1}University of Virginia\\
    \texttt{\{dpc3qt, ylkuo\}@virginia.edu}\\
}

\usepackage[T1]{fontenc}
\usepackage[utf8]{inputenc} % Usually already present
\usepackage{lmodern}

\begin{document}
\maketitle
\newcommand\dataset{\textsc{CaseEdit }}

%===============================================================================

\begin{abstract}
    Large language models (LLMs) exhibit strong performance on factual recall and general reasoning but struggle to adapt to user-specific, commonsense knowledge, a challenge particularly acute in small-parameter settings where computational efficiency is prioritized. We introduce \dataset, a new dataset and generation pipeline for evaluating localized, personalized commonsense knowledge editing in small LLMs to address this. Built upon the $ATOMIC_{20}^{20}$ commonsense graph, \dataset uses a multi-stage inference process to generate both typical and atypical contextual edits for household objects, paired with targeted evaluation questions across four axes: reliability, generalization, locality, and portability. We evaluate established knowledge editing methods using \dataset and demonstrate that AlphaEdit, a technique employing null-space projection to minimize interference with unrelated knowledge, consistently outperforms other methods when applied to an LLaMA 3.2 3B model, even in scalability tests, showing minimal ripple effects. Our results indicate that using \dataset with effective editing techniques like AlphaEdit allows small models to internalize high-quality, context-sensitive commonsense knowledge, paving the way for lightweight, personalized assistants.

    \begin{figure}[htbp]
      \centering
      \includegraphics[width=\textwidth]{images/editingvisual.png} % adjust width or use height
      \label{fig:my_label}
    \end{figure}
\end{abstract}

%===============================================================================

\section{Introduction}
	
    Large parameter language models, such as GPT-4o \cite{openai2024gpt4ocard} and LLaMA 3.1 405B \cite{grattafiori2024llama3herdmodels}, have demonstrated remarkable capabilities in handling complex queries, reasoning about abstract concepts, and adapting to nuanced contexts. Their extensive parameter count allows them to encode vast amounts of world knowledge and context, significantly enhancing their commonsense reasoning abilities. By drawing on richer representations and broader training datasets, large models excel at generalizing across diverse situations, making them well-suited for applications requiring nuanced understanding and inference \cite{zhang2024improvingdiversitycommonsensegeneration}. However, their significant computational and memory requirements make them impractical for edge computing applications. This is particularly relevant for personalized use cases, such as smart home assistants, where real-time adaptability, data privacy, and energy efficiency are critical. Smaller parameter models are ideal for such environments due to their lightweight architecture (See Appendix~\ref{appendix:parametercomparison} for more). \cite{tinychat2024}. However, they face unique challenges in commonsense reasoning, often falling short when tasked with adapting to highly personalized or context-specific requirements \cite{Li2021ASI}. For instance, common household objects are frequently repurposed in intuitive yet unconventional ways to meet the unique needs of individual households. A butter knife might serve as a makeshift screwdriver, or noise-canceling headphones might be used for sleeping rather than studying. Such adaptations arise not from randomness but from the specific habits, constraints, and preferences of each household. Similarly, context redefines assumptions; in a lactose-free household, for example, the term ``milk'' might intuitively refer to almond or oat milk rather than dairy milk. Both large and small models struggle to seamlessly integrate and adapt to personalized commonsense knowledge without explicit and repetitive prompting. Addressing this challenge requires frameworks and datasets designed specifically for commonsense knowledge editing, enabling models to intuitively reason about flexible, context-specific knowledge while preserving their broader functionality \cite{yao2023editinglargelanguagemodels,wang2024wise}.

We aim to bridge this gap by introducing \dataset and its creation framework. Designed to complement established knowledge editing techniques, this framework enables both large and small language models to adapt their internal representations to household-specific contexts. By facilitating the integration of intuitive, context-driven adaptations, this approach allows LLMs to function as more effective and personalized assistants, capable of reasoning flexibly about the dynamic and unique needs of individual households.

Our results when applying \dataset with AlphaEdit to an LLaMA 3.2 3B Instruct model were impressive. We found that the commonsense edits with AlphaEdit performed comparably to factual knowledge editing techniques in multiple-choice evaluations, despite commonsense editing being inherently more challenging due to its reliance on distributed knowledge representations across multiple layers, while factual editing typically requires minimal layer adjustments. Finally, we demonstrate that AlphaEdit, when paired with \dataset, effectively reduces the ripple effect.

%===============================================================================

\section{Related Work}

\subsection{Knowledge Editing}

Knowledge editing aims to efficiently modify the behavior of large language models (LLMs) to incorporate new information, correct inaccuracies, or customize responses without costly retraining \cite{zhang2024comprehensive}. Various approaches have been developed, broadly categorized into methods that directly modify model weights and those employing meta-learning or auxiliary networks.

Direct weight modification techniques often target specific layers identified as crucial for knowledge storage, typically within the transformer's feed-forward networks. \textbf{Rank-One Model Editing (ROME)} \cite{meng2022locating} and \textbf{Mass-Editing Memory in a Transformer (MEMIT)} \cite{meng2023masseditingmemorytransformer} are prominent examples. ROME applies rank-one updates to MLP weights to insert factual associations, located using causal tracing. MEMIT extends this concept, demonstrating scalability to thousands of simultaneous edits by precisely identifying and adjusting parameters, also leveraging causal mediation analysis to pinpoint knowledge storage locations \cite{wang2023easyedit}. While effective for factual updates, the distributed nature of commonsense knowledge poses challenges. \textbf{MEMIT-CSK} \cite{gupta2023editingcommonsensetransformers} adapts MEMIT specifically for commonsense, broadening the editing scope beyond subject tokens and using a moving average of causal effects across multiple layers to better capture distributed representations.

Another category involves meta-learning or using auxiliary models. \textbf{Model Editor Networks with Gradient Decomposition (MEND)} \cite{mitchell2021fast} trains a separate hypernetwork to predict parameter updates for specific edits. By learning a low-rank decomposition of the gradients associated with edits, MEND aims for efficient and generalizable updates without directly solving optimization problems for each edit instance.

Minimizing unintended side effects (``ripple effects'') is a key concern in knowledge editing. \textbf{AlphaEdit} \cite{fang2024alphaeditnullspaceconstrainedknowledge} addresses this by formulating the edit as a constrained optimization problem. It computes weight updates that satisfy the desired edit while explicitly minimizing changes in the null space of activations associated with preserved knowledge. This projection onto the null space helps localize the update and prevent interference with unrelated information, making it potentially suitable for complex or sequential edits, including those involving distributed commonsense knowledge.

\subsection{Commonsense Dataset}

Commonsense reasoning remains a significant area of focus for large language models (LLMs). Datasets capturing human-like commonsense are crucial resources for training and evaluating these models. The \textbf{ATOMIC$_{\mathbf{20}}^{\mathbf{20}}$} dataset is a prominent example, extending its predecessor, ATOMIC, by integrating symbolic and neural representations within a comprehensive neuro-symbolic knowledge graph \cite{sap2019atomicatlasmachinecommonsense, hwang2021comet}. Comprising 1.33 million tuples across 23 relation types (e.g., \texttt{ObjectUse}, \texttt{HinderedBy}), it covers social, physical, and event-centered reasoning. Curated via crowdsourcing to reflect human intuition, ATOMIC$_{\text{20}}^{\text{20}}$ provides structured knowledge that complements the implicit commonsense learned by LLMs during pre-training \cite{yao2023survey, sap2019atomicatlasmachinecommonsense, hwang2021comet}.

\begin{figure}[H]
    \centering
    \includegraphics[width=0.9\linewidth]{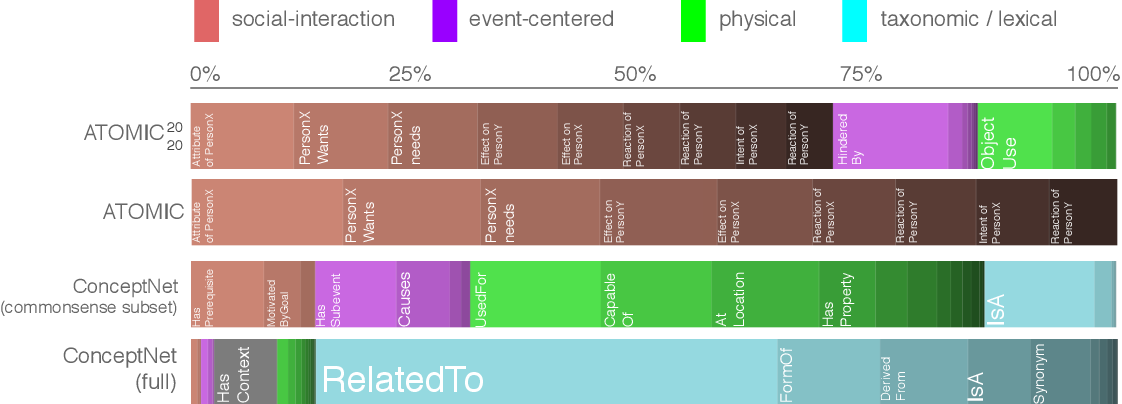}
    \vspace{0.5cm}
    \caption{\label{tab:atomic-fig}
    ATOMIC2020 tuple count distribution compared to other commonsense datasets \cite{hwang2021comet}}
\end{figure}

In LLM research, commonsense datasets like ATOMIC$_{\text{20}}^{\text{20}}$ and ConceptNet \cite{speer2017conceptnet55openmultilingual} are frequently used for various purposes. They serve as benchmarks for evaluating the reasoning capabilities of models \cite{li2021systematic}, provide data for fine-tuning models to enhance their commonsense understanding \cite{west2021symbolic}, or are integrated into retrieval-augmented generation (RAG) systems to ground model responses in explicit commonsense knowledge \cite{pan2023unifying}. These datasets help researchers probe the limits of LLMs and develop methods to improve their ability to reason about the everyday world.

Our work builds directly upon the foundation laid by ATOMIC$_{\text{20}}^{\text{20}}$. While existing uses often focus on evaluation or general fine-tuning, we adapt and extend the principles of ATOMIC$_{\text{20}}^{\text{20}}$ specifically for the context of \textit{knowledge editing}. We leverage its relational structures (\texttt{ObjectUse}, \texttt{HasProperty}, \texttt{AtLocation}) and subject matter as a basis for our \dataset generation pipeline (described in Section~\ref{sec:dataset_construction}). \dataset utilizes ATOMIC$_{\text{20}}^{\text{20}}$'s framework but focuses on generating paired typical and atypical scenarios for household objects, creating targeted edit examples and corresponding evaluation questions designed to assess the ability of knowledge editing techniques to instill personalized, context-dependent commonsense into smaller LLMs. This contrasts with using the dataset solely as a static benchmark, instead repurposing its structure to create dynamic editing tasks.

\section{Dataset Construction}
\subsection{Ground Truth Generation}
\label{sec:dataset_construction}
\dataset is designed to support the editing and evaluation of commonsense knowledge by curating typical and atypical contexts for household objects. The core components of this dataset include the subject, the target edit (representing the normal or typical understanding), the ground truth edit (capturing the atypical understanding), evaluation questions, and multiple choices tailored to assess the knowledge edits. We use the ATOMIC dataset as the foundational resource for \textbf{subject selection and relationship templates}. We generate edits that fall into three of the Physical-Entity Commonsense buckets: \textbf{ObjectUse}, which describes the everyday affordance or uses of objects; \textbf{HasProperty}, which denotes the relationship between an entity and its composition or characteristics; and \textbf{AtLocation}, a spatial relation that describes the location in/on/at which an entity is likely to be found. To generate atypical contexts, we employ GPT-4o-mini in a multi-step inference process (See Appendix~\ref{appendix:generation_prompts} for more details). First, for each selected subject, GPT-4o-mini is prompted to propose an atypical everyday household location for the object. For instance, while butter knives are typically associated with kitchens, GPT-4o-mini might suggest a garage as an unconventional location. In the second inference step, GPT-4o-mini generates an edit conditioned on the Physical-Entity Commonsense bucket and the atypical location. For example, if a butter knife is conditioned on ``ObjectUse'' and ``Garage'', the model might suggest new usage is ``tightening flathead screws.''

\begin{table}[H]
    \centering
    \footnotesize
    \renewcommand{\arraystretch}{1.3} % Reduce row height
    \setlength{\tabcolsep}{3pt} % Reduce column spacing
    \begin{tabular}{@{}p{1.8cm} p{2.2cm} p{2.2cm} p{4cm} p{3.3cm}@{}}
        \toprule
        \textbf{Subject} & \textbf{Plaus. Bucket} & \textbf{Target Edit} & \textbf{Unusual Everyday Location} & \textbf{New Ground Truth} \newline (Conditioned on unusual location) \\
        \midrule
        Butter Knife & ObjectUsage & Spreading & Garage Toolbox & Tightening flatheads \\
        \midrule
        Pillow & HasProperty & Soft and warm & Found inside a freezer & Cold and soothing \\
        \midrule
        Headphones & AtLocation & Study Room & Found on bedside table & Bedroom \\
        \bottomrule
    \end{tabular}
    \vspace{0.5cm}
    \caption{\label{tab:plausibility_shifts} Examples of \dataset knowledge editing chain creation pipeline. }
    \label{tab:table1}
\end{table}
\subsection{Evaluation Question Generation}
\label{sec:sec:dataset_construction}
In the next stage, an additional inference loop with GPT-4o-mini is used to generate evaluation questions designed to evaluate the knowledge edits against four key metrics: \textbf{reliability}, \textbf{generalization}, \textbf{locality}, and \textbf{portability} \cite{zhang2024comprehensivestudyknowledgeediting}. The performance of the model is evaluated using a multiple-choice question format (MCQ) to ensure reproducibility and a systematic evaluation process. In this framework, the previous ground truth, the newly generated ground truth, and three unrelated distractor choices are curated and randomly assigned labels from \textbf{A} to \textbf{E}. This structured approach enables a robust and standardized assessment of the model's capacity to produce accurate and contextually appropriate knowledge edits. As illustrated in Table~\ref{tab:table1}, the ground truth generation involves two sequential inference steps. First, an LLM call takes the subject, plausibility bucket, and original target edit as input to generate an atypical everyday location (refer to Appendix~\ref{appendix:generation_prompts} for example prompts). Second, another inference call uses the subject, plausibility bucket, target edit, and the generated unusual location to produce the new ground truth statement. Including the original target edit in this second step guides the LLM to generate outputs distinct from the initial ground truth. Subsequently, as shown in Table~\ref{tab:table2}, a final inference stage generates the four evaluation questions. This stage takes the subject and the newly generated ground truth as input (see Appendix~\ref{appendix:generation_prompts} for prompts).

\begin{table}[H]
    \centering
    \footnotesize
    \renewcommand{\arraystretch}{1.3} % Reduce row height
    \setlength{\tabcolsep}{5pt} % Reduce column spacing
    \begin{tabular}{@{}p{1.8cm} p{2.2cm} p{2.2cm} p{4cm} p{3.3cm}@{}}
        \toprule
        \textbf{Subject} & \textbf{Reliability} & \textbf{Generalization} & \textbf{Locality} & \textbf{Portability}  \\
        \midrule
        Butter Knife & What do I use my \textcolor{blue}{butter knife} for? & What can I use to \textcolor{blue}{tighten a screw}? & What is a \textcolor{red}{chef's knife} used for? & I lost my \textcolor{blue}{butter knife}, what can I use instead? \\
        \midrule
        Pillow & What are the characteristics of a \textcolor{blue}{pillow}? & What is \textcolor{blue}{cold and soothing}? & What are the characteristics of a \textcolor{red}{mattress}? & How does a \textcolor{blue}{pillow} help with headaches? \\
        \midrule
        Headphones & Where are \textcolor{blue}{headphones} used? & What might be on my \textcolor{blue}{bedside table} & Where could I find my \textcolor{red}{smartphone} & Why are \textcolor{blue}{headphones} good to have? \\
        \bottomrule
    \end{tabular}
    \vspace{0.5cm}
    \caption{Examples of \dataset evaluation questions. Tokens activating the edited layer are highlighted in blue, while potentially entangled tokens that should remain unchanged are highlighted in red.}
    \label{tab:table2}
\end{table}

\begin{table}[h]
    \centering
    \small
    \setlength{\tabcolsep}{5pt}
    \renewcommand{\arraystretch}{1.2}
    
    \label{tab:dataset_stats}
    \begin{tabular}{lc}
        \hline
        \textbf{Statistic} & \textbf{Value} \\
        \hline
        Number of Subject Edits & 900 \\
        Total Eval. Questions & 3,600 \\
        Total MCQ Choices & 18,000 \\
        Avg. Subject Tokens & 2.4 \\
        Avg. New Ground Truth Tokens& 3.6 \\
        Avg. Previous Truth Tokens & 3.2 \\
        Avg. Evaluation Question Tokens & 10.3 \\
        Avg. MCQ Choice Tokens & 2.7 \\
        \hline
    \end{tabular}
    \vspace{0.5cm}
    \caption{\dataset Statistics}
\end{table}

\section{Experiment}
\label{sec:experiments} % Optional: Add a label for cross-referencing

We evaluate the effectiveness of various knowledge editing techniques on our generated dataset, \dataset. This section details the models, evaluation metrics, and experimental procedures used.

\subsection{Models and Editing Setups}
\label{subsec:models_setups} % Optional: Add a label

Our experiments primarily utilize the \textbf{LLaMA 3.2 3B-Instruct} model as the base model for evaluating AlphaEdit, ROME, MEND, and MEMIT \cite{fang2024alphaeditnullspaceconstrainedknowledge, meng2022locating, mitchell2021fast, meng2023masseditingmemorytransformer}. This model provides a strong baseline for assessing editing performance on a contemporary, instruction-following architecture.

For the \textbf{MEMIT-CSK} \cite{gupta2023editingcommonsensetransformers} evaluation, we use the \textbf{GPT-2 XL} (1.5B parameters) model. This choice aligns with the experimental setup often used in the original MEMIT-CSK research and allows for comparison within the context of its typical evaluation framework.

All editing methods were implemented using standard configurations, often relying on default hyperparameters provided by common knowledge editing libraries (e.g., EasyEdit \cite{wang2023easyedit}) unless otherwise specified in the respective method's original publication. Specific layer targeting and other method-specific parameters followed the recommendations from their source papers. For instance, MEMIT and ROME edits target specific MLP layers identified via causal tracing or related analyses, while AlphaEdit computes updates based on its null-space projection constraint.

\subsection{Evaluation Metrics}
\label{subsec:metrics} % Optional: Add a label

To comprehensively assess the performance of knowledge editing methods, we adopt four standard metrics widely used in the literature \cite{zhang2024comprehensive}:

\begin{itemize}
    \item \textbf{Reliability:} Measures whether the edit successfully modifies the model's output for the specific input example provided during the edit. It assesses the direct success of the update ($f_{\theta_e}(x_e) = y_e$).
    \item \textbf{Generalization:} Evaluates if the model correctly applies the edited knowledge to semantically similar inputs or paraphrased versions of the original edit query. This tests if the edit extends appropriately within its intended scope.
    \item \textbf{Locality:} Assesses whether the edit unintentionally alters the model's predictions on unrelated inputs that should not be affected by the specific knowledge update. High locality indicates minimal negative side effects or "ripple effects" on the model's broader knowledge.
    \item \textbf{Portability:} Measures if the newly acquired knowledge through the edit can be correctly applied in more complex, multi-hop reasoning scenarios or downstream tasks that logically depend on the edited fact or concept.
\end{itemize}

These metrics allow us to evaluate not only if an edit is successful (Reliability) but also if it behaves as expected within its intended scope (Generalization), avoids damaging other knowledge (Locality), and integrates usefully into broader reasoning (Portability).

\subsection{Experimental Setup}
\label{subsec:setup} % Optional: Add a label

We conduct two main experiments to evaluate the performance of the selected knowledge editing techniques using the metrics defined above:

\begin{itemize}
    \item \textbf{Fixed Edits Test:} We randomly select 50 subjects from \dataset. For each subject, the corresponding edit (changing from the typical to the atypical context) is applied using each knowledge editing technique (AlphaEdit, ROME, MEND, MEMIT on LLaMA 3 8B; MEMIT-CSK on GPT-2 XL). Edits are applied \textit{sequentially} to the model, meaning each subsequent edit modifies the model state resulting from the previous edit. This sequential application allows us to evaluate performance under cumulative modifications and assess the potential for interference or compounding ripple effects between edits. After applying all 50 sequential edits, we evaluate the model's performance on the evaluation questions associated with these edits across the four metrics.
    \item \textbf{Scalability Test:} To assess performance under increasing edit load, we vary the number of edits ($n$) applied to the base model across multiple levels ($n = 10, 20, 50, 100, 200$). For each level of $n$, we randomly select $n$ distinct edits from \dataset and apply them \textit{sequentially} to the appropriate base model for each technique. This setup explicitly tests how the editing methods handle increasing numbers of potentially interfering updates. We then evaluate the model's performance on the corresponding evaluation questions for those $n$ edits.
\end{itemize}

For both tests, we employ the multiple-choice question (MCQ) format defined in our \dataset generation process to systematically evaluate the model outputs against the four metrics. We also analyze the model's confidence by examining the softmax probability distribution over the MCQ answer choices (A-E) before and after edits, providing insight into how certainty shifts with knowledge modification (see Appendix~\ref{appendix:confidence_analysis} for details). 

\section{Results}

\subsection{AlphaEdit outperforms other editing techniques}
\begin{figure}[H]
    \centering
    \begin{minipage}[b]{0.45\linewidth}
        \centering
        \includegraphics[width=\linewidth]{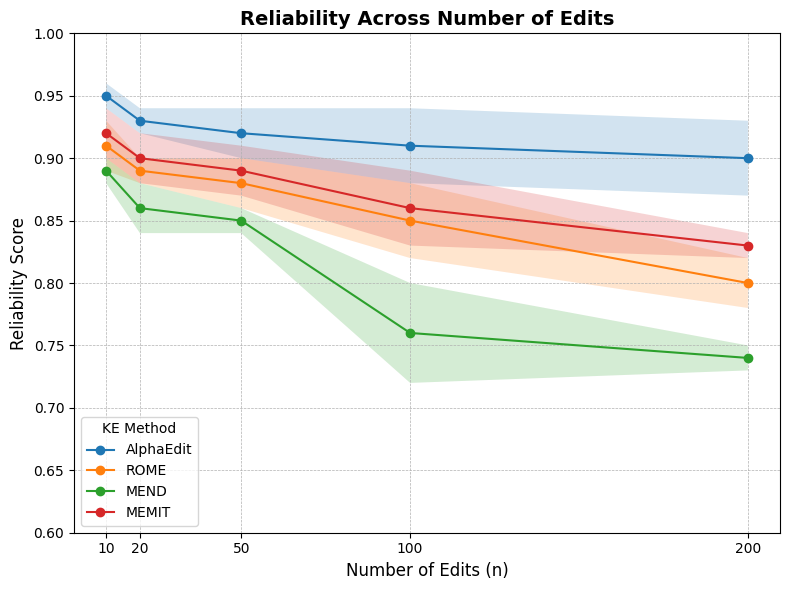}
    \end{minipage}
    \hfill
    \begin{minipage}[b]{0.45\linewidth}
        \centering
        \includegraphics[width=\linewidth]{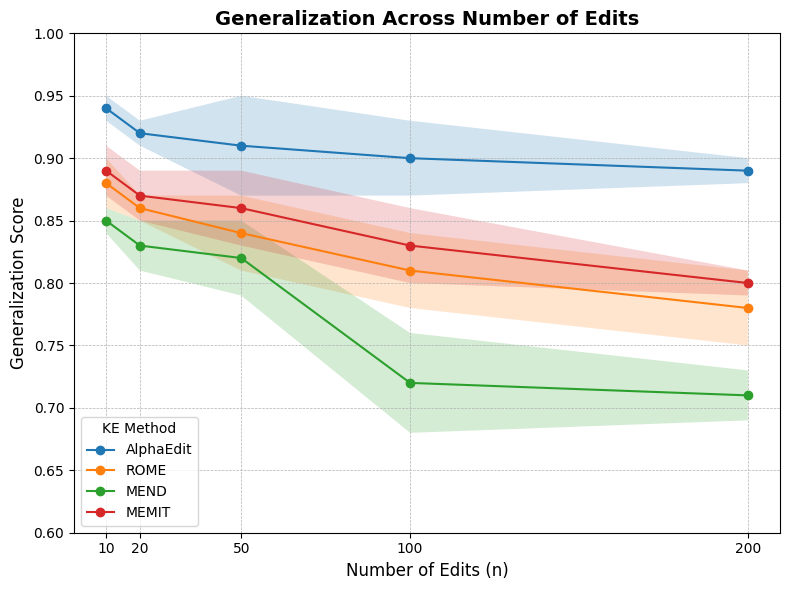}
    \end{minipage}
    
    \vspace{0.5cm}

    \begin{minipage}[b]{0.45\linewidth}
        \centering
        \includegraphics[width=\linewidth]{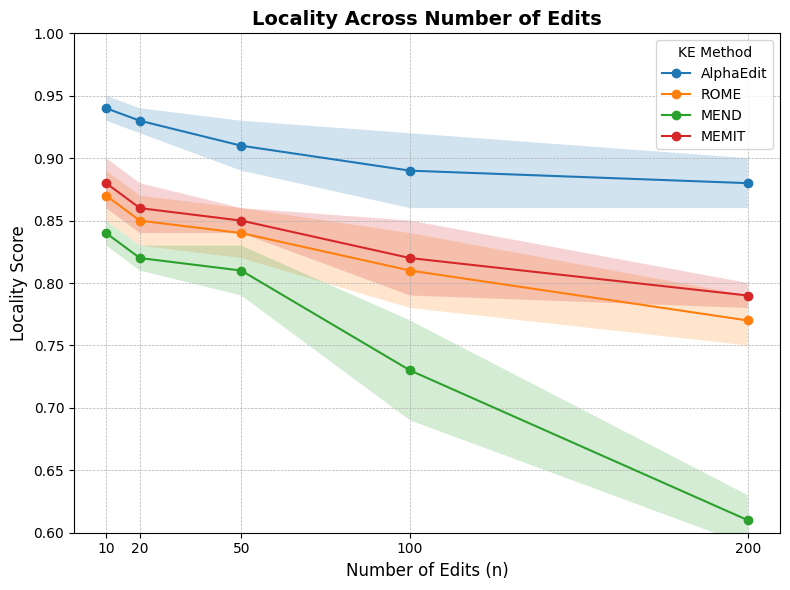}
    \end{minipage}
    \hfill
    \begin{minipage}[b]{0.45\linewidth}
        \centering
        \includegraphics[width=\linewidth]{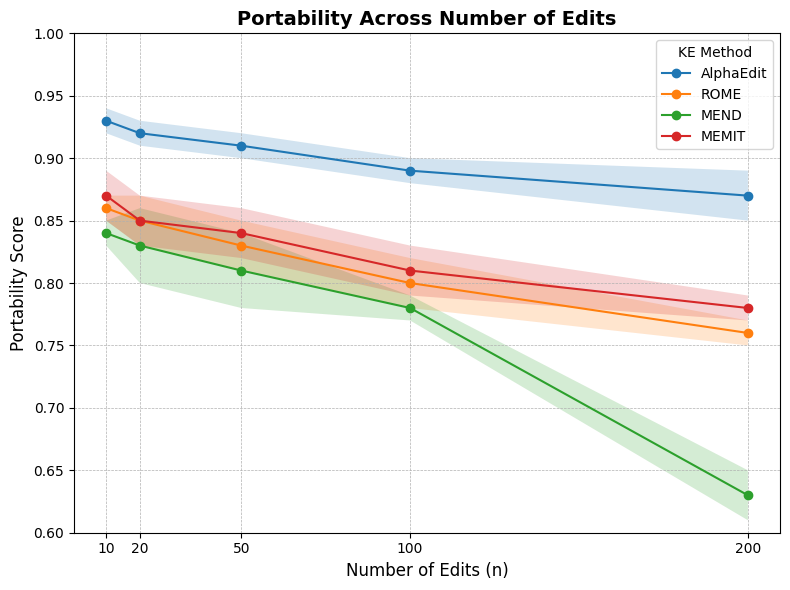}
    \end{minipage}

    \caption{Changes in model reliability, generalization, locality, and portability over the number of commonsense edits.}
    \label{fig:commonsense-metrics}
\end{figure}

Based on the results presented in Table~\ref{tab:edit-results}, AlphaEdit outperforms all other knowledge editing methods on \dataset across all evaluated metrics on our dataset. For the scalability test, as seen in Figure~\ref{fig:commonsense-metrics}, we observe that AlphaEdit is less resistant to an increasing number of commonsense edits compared to other editing methods. \footnote{MEMIT-CSK uses a smaller and older GPT-2XL, which attributes to the relatively poor performance}

\begin{table}[htbp] % Consider [htbp] instead of [H] for better float placement
  \centering
  \caption{Performance Metrics Across Editing Techniques (n=50)} % Added caption
  \label{tab:edit-results} % Label for referencing the table
  \begin{tabular}{lcccc}
    \toprule % Use booktabs rule
    \textbf{Technique}  & \textbf{Reliability} & \textbf{Generalization} & \textbf{Locality} & \textbf{Portability} \\
    \midrule % Use booktabs rule
    Base Model    & 0.00 $\pm$ 0.00 & 0.00 $\pm$ 0.00 & 0.00 $\pm$ 0.00 & 0.00 $\pm$ 0.00 \\
    AlphaEdit  & \textbf{0.93} $\pm$\textbf{ 0.02} & \textbf{0.91} $\pm$ \textbf{0.02} & \textbf{0.87} $\pm$ \textbf{0.02} & \textbf{0.90} $\pm$ \textbf{0.01} \\
    MEMIT-CSK    & 0.87 $\pm$ 0.02 & 0.83 $\pm$ 0.01 & 0.81 $\pm$ 0.02 & 0.84 $\pm$ 0.02 \\
    ROME         & 0.88 $\pm$ 0.02 & 0.84 $\pm$ 0.03 & 0.82 $\pm$ 0.02 & 0.80 $\pm$ 0.02 \\
    MEND         & 0.86 $\pm$ 0.01 & 0.81 $\pm$ 0.03 & 0.78 $\pm$ 0.02 & 0.76 $\pm$ 0.03 \\
    MEMIT        & 0.90 $\pm$ 0.02 & 0.87 $\pm$ 0.03 & 0.86 $\pm$ 0.01 & 0.85 $\pm$ 0.02 \\
    \bottomrule % Use booktabs rule
  \end{tabular}
\end{table}

\subsection{Models Exhibit Increased Uncertainty with Scaled Edits}

\begin{figure}[H]
    \centering
    \begin{minipage}[b]{0.45\linewidth}
        \centering
        \includegraphics[width=\linewidth]{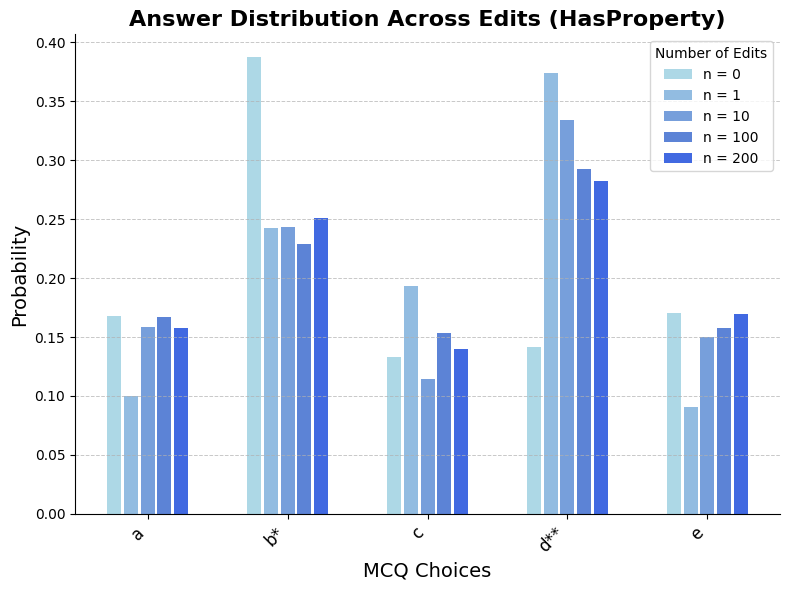}
        \small (a) HasProperty
    \end{minipage}
    \hfill
    \begin{minipage}[b]{0.45\linewidth}
        \centering
        \includegraphics[width=\linewidth]{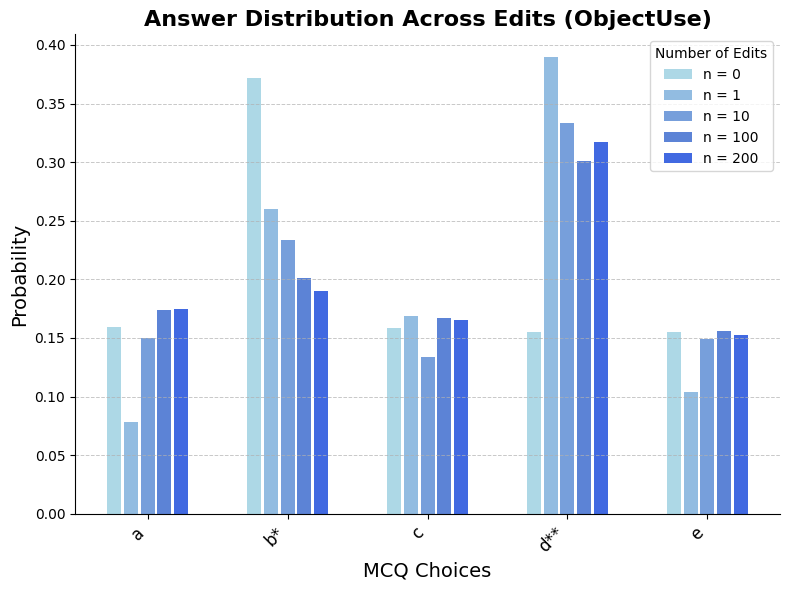}
        \small (b) ObjectUse
    \end{minipage}

    \vspace{0.5cm}

    \begin{minipage}[b]{0.45\linewidth}
        \centering
        \includegraphics[width=\linewidth]{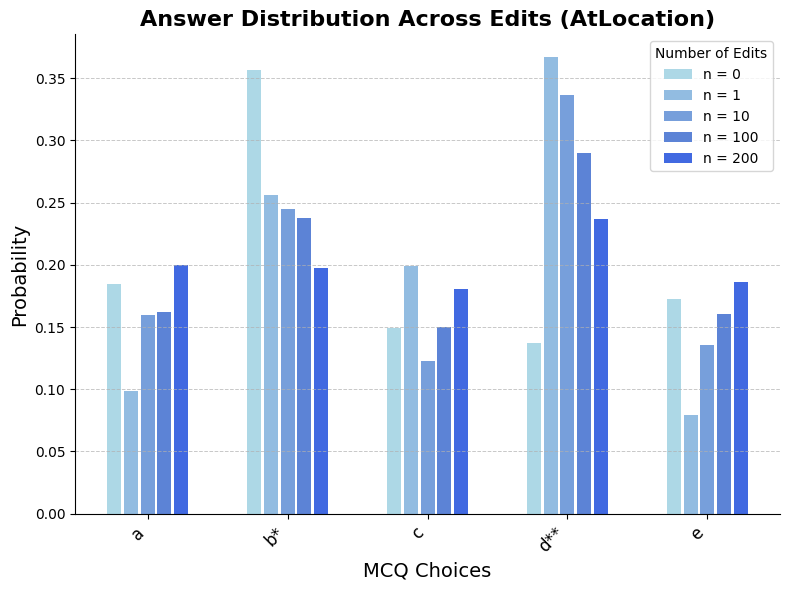}
        \small (c) AtLocation
    \end{minipage}

    \caption{Next-token probabilities and entropy during MCQ evaluations across relational buckets: HasProperty, ObjectUse, and AtLocation.}
    \label{fig:logit-change}
\end{figure}

As depicted in Figure~\ref{fig:logit-change}, we analyzed the model's confidence by examining the probability distribution over the five multiple-choice answers. This distribution is obtained by applying the softmax function to the output logits generated by the model for each MCQ choice (see Appendix~\ref{appendix:confidence_analysis} for details). The analysis presented here focuses specifically on edits performed using the \textbf{AlphaEdit} method.

To facilitate comparison, we standardized the plotting such that the previous truth corresponds to option B and the new ground truth (the target of the edit) corresponds to option D, although the choices were randomized during actual evaluation. Before the edit (n=0), the model typically shows high confidence (low uncertainty) in the original truth (option B). After applying a single AlphaEdit commonsense knowledge edit (n=1), the probability mass shifts significantly towards the new ground truth (option D). However, the model often retains some residual probability for the original truth (B) and distributes the remaining probability among the distractors (A, C, E), indicating the edit was successful but introduced some uncertainty.

As we scaled the number of additional, unrelated sequential edits (n=10,20,...200), we observed a gradual decrease in the probability assigned to the correct new ground truth (D) and a slight increase in probabilities for other options. This suggests that the model's confidence in the specific edited fact decreases (i.e., uncertainty increases) as more potentially interfering edits accumulate. This indicates that while the ripple effect from sequential edits impacts certainty, it is not substantial enough, at least with AlphaEdit within this scale, to completely override the correction.

We hypothesize that the degree of this uncertainty increase is related to the editing mechanism's ability to localize updates. Methods like AlphaEdit, which employ techniques such as null-space projection to perform cleaner edits and minimize interference with unrelated knowledge, likely mitigate this effect more effectively than methods with less constrained update procedures. Consequently, the observed increase in uncertainty might be less pronounced with AlphaEdit due to its reduced ripple effects compared to what is seen with other techniques under similar sequential editing conditions (See Appendix~\ref{sec:llm-confidence-more}).

%===============================================================================

\section{Conclusion}
\label{sec:conclusion}

In this work, we introduce \dataset and a novel dataset generation framework designed to produce variants of commonsense knowledge editing datasets. The effectiveness of \dataset was demonstrated through experiments that implemented established knowledge editing methods on a small-parameter LLaMA 3.2 3B model. Notably, AlphaEdit, a generalized factual knowledge editing method, achieved performance comparable to MEMIT-CSK, a commonsense-specific editing approach, on \dataset, highlighting the viability of our framework. More broadly, \dataset establishes that a multistage inference-based data generation pipeline offers a promising avenue for modeling the inherently human-like reasoning demands of commonsense knowledge. This capability to effectively edit commonsense knowledge opens the path to creating highly personalized and adaptable small-parameter LLMs. Such models hold significant potential for deployment in edge computing environments, enabling user-specific customization of household AI assistants to better reflect preferences and contextual nuances.

%===============================================================================

\section{Limitations}

Systematic evaluation of \dataset was limited by the lack of large-scale human evaluations. Future work should incorporate crowd-sourced evaluations using Amazon Mechanical Turk to assess edit plausibility. An RLHF pipeline could further refine the edits and improve model alignment \cite{ouyang2022traininglanguagemodelsfollow}. Compute and time constraints restricted the number of edits studied. Expanding evaluations to a larger set of edits would provide insights into the scalability of knowledge editing techniques. RAG and fine-tuning are alternative methods for updating and retrieving knowledge \cite{lewis2021retrievalaugmentedgenerationknowledgeintensivenlp}. Comparing these approaches to knowledge editing techniques such as AlphaEdit on the same dataset would highlight their respective advantages and limitations. Further research should investigate the impact of model size on knowledge editing performance, particularly in terms of edit stability and unintended generalization effects. Evaluating knowledge editing on LLMs trained for reasoning, such as DeepSeek-r1 \cite{deepseekai2025deepseekr1incentivizingreasoningcapability}, would determine whether certain architectures or training methodologies improve commonsense reasoning and edit robustness. Additionally, exploring whether the single-line edit vector injection mechanism from AlphaEdit---specifically, the modified forward pass using \texttt{hidden\_states~+=~edit\_vector}---can be integrated into the Moving AIE architecture of MEMIT-CSK may lead to a hybrid model with improved commonsense editing performance. This hybridization could leverage AlphaEdit's simplicity and locality with MEMIT-CSK’s structured edit propagation, potentially enhancing edit fidelity while maintaining generalization control.

\newpage
\section{Appendix}
\label{sec:appendix}

\subsection{LLM Parameter Data}
\begin{figure}[H]
    \centering
    \includegraphics[width=\linewidth]{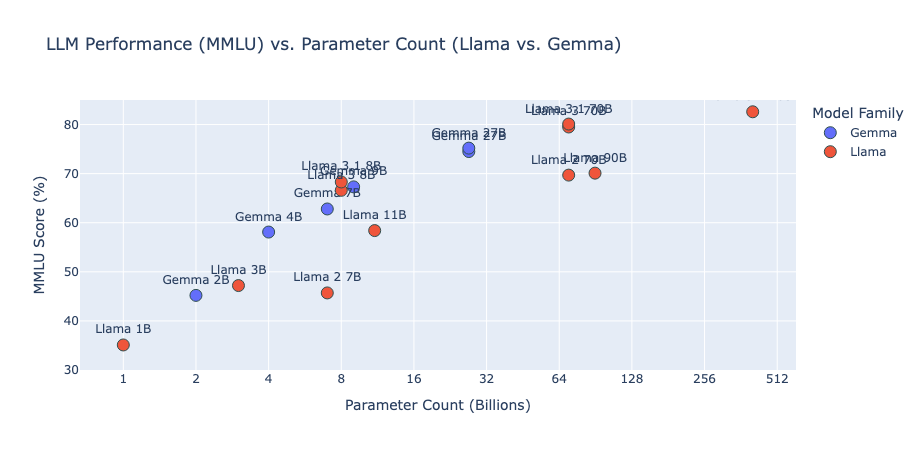}
    \caption{\label{appendix:parametercomparison} MMLU benchmark performance across different parameter sizes for Llama and Gemma models. Larger models generally yield higher scores.
}
\end{figure}

\subsection{Mathematical Derivations of Knowledge Editing Methods}

\subsubsection{Rank-One Model Editing (ROME)}

ROME \cite{meng2022locating} introduces a method to edit factual associations in transformer-based language models by performing a rank-one update to the weights of a specific feed-forward network (FFN) layer. The approach treats the FFN as a key-value memory, where the key represents the subject and the value represents the associated information.

Given:
\begin{itemize}
    \item A subject representation vector $\mathbf{k} \in \mathbb{R}^d$ (key),
    \item A desired new value vector $\mathbf{v} \in \mathbb{R}^d$,
    \item The original weight matrix $W \in \mathbb{R}^{d \times d}$ of the FFN layer.
\end{itemize}

ROME computes an update $\Delta W$ to the weight matrix as:
\[
\Delta W = (\mathbf{v} - W \mathbf{k}) \mathbf{k}^\top / (\mathbf{k}^\top \mathbf{k})
\]

The updated weight matrix becomes:
\[
W' = W + \Delta W
\]

This update ensures that the FFN maps the key $\mathbf{k}$ to the new value $\mathbf{v}$, effectively altering the model's response to inputs related to the subject.

\subsubsection{Mass-Editing Memory in a Transformer (MEMIT)}

MEMIT \cite{meng2022masseditingmemorytransformer} extends the ROME approach to handle batch editing of multiple facts simultaneously. It distributes the updates across multiple layers to maintain model stability and performance.

Given a set of $n$ edits, each consisting of:
\begin{itemize}
    \item A key vector $\mathbf{k}_i \in \mathbb{R}^d$,
    \item A desired value vector $\mathbf{v}_i \in \mathbb{R}^d$,
\end{itemize}
for $i = 1, \ldots, n$, MEMIT aims to find weight updates $\Delta W_l$ for each layer $l$ such that:
\[
\mathbf{v}_i = \left( W_l + \Delta W_l \right) \mathbf{k}_i, \quad \forall i
\]

To solve this, MEMIT formulates a least-squares problem:
\[
\min_{\{\Delta W_l\}} \sum_{i=1}^n \left\| \mathbf{v}_i - \left( W_l + \Delta W_l \right) \mathbf{k}_i \right\|^2 + \lambda \left\| \Delta W_l \right\|_F^2
\]

where $\lambda$ is a regularization parameter, and $\|\cdot\|_F$ denotes the Frobenius norm. The solution involves computing the optimal $\Delta W_l$ that minimizes the reconstruction error while keeping the updates small to preserve the model's original behavior.

\subsubsection{Model Editor Networks with Gradient Decomposition (MEND)}

MEND \cite{mitchell2022fast} introduces a meta-learning approach to perform rapid and localized edits to a language model using gradient information. It employs a hypernetwork to predict low-rank updates to the model's parameters based on the gradient of a loss function computed from a single edit example.

Given:
\begin{itemize}
    \item A pre-trained model with parameters $\theta$,
    \item A loss function $\mathcal{L}(\theta)$ computed on the edit example,
    \item The gradient $\nabla_\theta \mathcal{L}(\theta)$,
\end{itemize}
MEND computes a low-rank decomposition of the gradient:
\[
\nabla_\theta \mathcal{L}(\theta) \approx U V^\top
\]

where $U \in \mathbb{R}^{d \times r}$ and $V \in \mathbb{R}^{d \times r}$ with $r \ll d$. A hypernetwork $H$ is trained to predict the update $\Delta \theta$ as:
\[
\Delta \theta = H(U, V)
\]

The model parameters are then updated as:
\[
\theta' = \theta + \Delta \theta
\]

This approach allows for efficient and scalable edits by leveraging the structure of the gradient and the predictive capabilities of the hypernetwork.

\label{appendix:editingmath}

\subsection{Generation Prompts}
\label{appendix:generation_prompts}

This section details the prompts used with the GPT-4o-mini model via its API for the different stages of the CASEEDIT dataset generation.

\subsubsection{Inference Step 1: Generate Unusual Everyday Location}
\label{appendix:prompt_step1}

This prompt generates an atypical, yet plausible, household location for a given object.

\begin{lstlisting}[style=pythonstyle, caption={API Call for Unusual Location Generation}, label={lst:prompt1}]
import openai


# --- Placeholder variables ---
# subject = "Butter Knife"
# plausibility_bucket = "ObjectUse"
# target_edit = "Spreading and cutting"
# ---------------------------

response = openai.chat.completions.create(
    model="gpt-4o-mini",
    messages=[
        {"role": "system",
         "content": "You are an expert in commonsense reasoning. Your task is to propose an unusual, yet plausible, everyday household location for a common object, given its typical use or property. The location should be different from where the object is normally found but still conceivable within a home environment. Output only the location name (e.g., 'Garage Toolbox', 'Bathroom Cabinet', 'Under the Bed')."
        },
        {"role": "user",
         "content": f"Generate an unusual everyday household location based on the following:\n\nSubject: {subject}\nPlausibility Bucket: {plausibility_bucket}\nTypical Use/Property (Target Edit): {target_edit}\n\nUnusual Everyday Household Location:"
        }
    ],
    temperature=0.7,
    max_tokens=15
)

unusual_location = response.choices[0].message.content
print(f"Generated Unusual Location: {unusual_location}")
# Expected Example Output: Garage Toolbox
\end{lstlisting}

\subsubsection{Inference Step 2: Generate New Ground Truth}
\label{appendix:prompt_step2}

This prompt generates a new plausible use or property for the object based on the unusual location generated in Step 1.

\begin{lstlisting}[style=pythonstyle, caption={API Call for New Ground Truth Generation}, label={lst:prompt2}]

# --- Placeholder variables ---
# subject = "Butter Knife"
# plausibility_bucket = "ObjectUse"
# target_edit = "Spreading and cutting"
# unusual_location = "Garage Toolbox" # From previous step
# ---------------------------

response = openai.chat.completions.create(
    model="gpt-4o-mini",
    messages=[
        {"role": "system",
         "content": "You are an expert in commonsense reasoning. Given an object, its typical use/property, a related commonsense category (Plausibility Bucket), and an unusual household location where it might be found, generate a plausible new use or property for the object specifically related to that unusual location. The new use/property should be distinct from the typical one provided. Output only the new use or property statement (e.g., 'Tightening flatheads', 'Cold and soothing', 'Storing small screws')."
        },
        {"role": "user",
         "content": f"Generate a new ground truth statement based on the following:\n\nSubject: {subject}\nPlausibility Bucket: {plausibility_bucket}\nTypical Use/Property (Target Edit): {target_edit}\nUnusual Everyday Household Location: {unusual_location}\n\nNew Ground Truth (related to the unusual location and distinct from typical use/property):"
        }
    ],
    temperature=0.7,
    max_tokens=25
)

new_ground_truth = response.choices[0].message.content
print(f"Generated New Ground Truth: {new_ground_truth}")
# Expected Example Output: Tightening flatheads
\end{lstlisting}

\subsubsection{Inference Step 3: Generate Evaluation Questions}
\label{appendix:prompt_step3}

This prompt generates four distinct evaluation questions based on the subject and the new ground truth from Step 2.

\begin{lstlisting}[style=pythonstyle, caption={API Call for Evaluation Question Generation}, label={lst:prompt3}]


# --- Placeholder variables ---
# subject = "Butter Knife"
# new_ground_truth = "Tightening flatheads" # From previous step
# ---------------------------

response = openai.chat.completions.create(
    model="gpt-4o-mini",
    messages=[
        {"role": "system",
         "content": "You are an expert in evaluating knowledge editing in language models. Given an object (Subject) and a newly established atypical commonsense fact about it (New Ground Truth), generate four distinct evaluation questions designed to test different aspects of knowledge editing:\n1.  **Reliability:** Directly ask about the New Ground Truth.\n2.  **Generalization:** Ask a question that requires applying the New Ground Truth to a slightly different but related concept or phrasing.\n3.  **Locality:** Ask about a related but distinct object or concept that should *not* have been affected by the edit.\n4.  **Portability:** Ask a question that requires using the New Ground Truth in a simple reasoning step or application context.\n\nFormat the output as a JSON object with keys 'Reliability', 'Generalization', 'Locality', 'Portability' and the corresponding questions as string values."
        },
        {"role": "user",
         "content": f"Generate four evaluation questions based on the following:\n\nSubject: {subject}\nNew Ground Truth: {new_ground_truth}\n\nOutput:"
        }
    ],
    temperature=0.6,
    max_tokens=200,
    response_format={"type": "json_object"} # Request JSON output
)

# Assuming the response content is a JSON string
evaluation_questions = json.loads(response.choices[0].message.content)
print("Generated Evaluation Questions:")
print(json.dumps(evaluation_questions, indent=2))
# Expected Example Output (JSON structure):
# {
#   "Reliability": "What do I use my butter knife for now?",
#   "Generalization": "What household item can I use to tighten a flathead screw?",
#   "Locality": "What is a Phillips head screwdriver used for?",
#   "Portability": "I lost my screwdriver, what could I use from the toolbox instead to tighten a loose flathead screw on the shelf?"
# }
\end{lstlisting}

\subsection{Confidence Analysis via Softmax Probabilities}
\label{appendix:confidence_analysis}

In our multiple-choice question (MCQ) evaluations, the language model generates logits for each potential answer choice (A, B, C, D, E). Logits are the raw, unnormalized scores output by the final layer of the model before any activation function is applied. To interpret these scores as probabilities and analyze the model's confidence or uncertainty regarding the correct answer, we apply the softmax function.

Let $z = (z_A, z_B, z_C, z_D, z_E)$ be the vector of logits produced by the model for the five answer choices corresponding to a specific evaluation question. The softmax function converts this vector into a probability distribution $p = (p_A, p_B, p_C, p_D, p_E)$, where each $p_i$ represents the model's estimated probability for choice $i$, and $\sum_{i \in \{A,B,C,D,E\}} p_i = 1$. The probability for a specific choice $i$ is calculated as:

\[
p_i = \frac{e^{z_i}}{\sum_{j \in \{A,B,C,D,E\}} e^{z_j}}
\]

This probability distribution $p$ reflects the model's confidence distribution across the available choices. A distribution heavily skewed towards one choice indicates high confidence, while a more uniform distribution suggests higher uncertainty.

To quantify this uncertainty, we calculate the Shannon entropy $H(p)$ of the probability distribution:

\[
H(p) = - \sum_{i \in \{A,B,C,D,E\}} p_i \log_2(p_i)
\]

where, by convention, $0 \log_2(0) = 0$.

The entropy $H(p)$ measures the average amount of information or "surprise" inherent in the distribution.
\begin{itemize}
    \item \textbf{Maximum Entropy:} Occurs when the distribution is uniform ($p_A = p_B = ... = p_E = 1/5$), indicating maximum uncertainty as the model assigns equal probability to all choices. In this case, $H(p) = \log_2(5) \approx 2.32$ bits.
    \item \textbf{Minimum Entropy:} Occurs when the model assigns probability 1 to a single choice and 0 to all others (e.g., $p_D = 1, p_{i \neq D} = 0$), indicating maximum confidence or certainty. In this case, $H(p) = 0$ bits.
\end{itemize}

By analyzing the probability distributions (as visualized in Figure~\ref{fig:logit-change} in the main text) and their corresponding entropy values before and after edits, and as the number of edits increases, we gain insight into how knowledge editing impacts the model's certainty about both the original and the newly edited information, as well as its susceptibility to interference from unrelated edits. Lower entropy post-edit for the target answer generally indicates a more confident and successful edit, while increasing entropy with more edits can signal degradation or interference.

\subsection{LLM Confidence with Alternative Editing Methods}
\label{sec:llm-confidence-more}
\begin{table}[htbp] % Use placement specifiers like [htbp] for flexibility
    \centering
    \caption{New Ground Truth Probability (D) Across Edits}
    \label{tab:prob_d_across_edits}
    \begin{tabular}{lcccc} % Changed lccccc to lcccc
        \toprule
        Editing Method & \multicolumn{4}{c}{Number of Sequential Edits (n)} \\ % Changed \multicolumn{5}{c} to \multicolumn{4}{c}
        \cmidrule(lr){2-5} % Changed \cmidrule(lr){2-6} to \cmidrule(lr){2-5}
                       & n=0  & n=1  & n=10 & n=100 \\ % Removed n=200
        \midrule
    
        AlphaEdit      & 0.14 & 0.37 & 0.33 & 0.29  \\ % Removed last value (0.28)
        MEMIT          & 0.14  & 0.40  & 0.31  & 0.22  \\ % Removed last value
        ROME           & 0.14  & 0.33  & 0.25  & 0.23   \\ % Removed last value
        MEND           & 0.14  & 0.34  & 0.24  & 0.23 \\ % Removed last value
        % MEMIT-CSK    &   & ...  & ...  & ...   \\ 
        \bottomrule
    \end{tabular}
    \\ \vspace{0.5em}
    \footnotesize
    \textit{Note:} The n=0 column represents the base model's probability for the eventual 'new ground truth' choice before any edits. 
\end{table}

%===============================================================================
\newpage
\bibliography{example}  % .bib

\end{document}